\documentclass[10pt,conference]{IEEEtran}
\IEEEoverridecommandlockouts
\usepackage{cite}
\usepackage{amsmath,amssymb,amsfonts}
\usepackage{algorithmic}
\usepackage{graphicx}
\usepackage{textcomp}
\usepackage{xcolor}
\usepackage{multirow}
\usepackage{comment}
\usepackage{caption}
\usepackage{tabularx}
\def\BibTeX{{\rm B\kern-.05em{\sc i\kern-.025em b}\kern-.08em
    T\kern-.1667em\lower.7ex\hbox{E}\kern-.125emX}}
    
\usepackage{tikz}
\usepackage{textcomp}
\usepackage[doipre={DOI:~}]{uri}
\usepackage{lipsum}
\newcommand\copyrighttext{%
  \footnotesize \textcopyright 2022 IEEE. Personal use of this material is permitted.  Permission from IEEE must be obtained for all other uses, in any current or future media, including reprinting/republishing this material for advertising or promotional purposes, creating new collective works, for resale or redistribution to servers or lists, or reuse of any copyrighted component of this work in other works. 
  
  Accepted as a conference paper at the 2022 IEEE International Symposium on Circuits and Systems (ISCAS).}
\newcommand{\copyrightnotice}{%
\begin{tikzpicture}[remember picture,overlay]
\node[anchor=south,yshift=10pt] at (current page.south) {\fbox{\parbox{\dimexpr\textwidth-\fboxsep-\fboxrule\relax}{\copyrighttext}}};
\end{tikzpicture}%
}

\begin{document}

\title{C-NMT: A Collaborative Inference Framework for Neural Machine Translation}

\author{\IEEEauthorblockN{Yukai Chen\IEEEauthorrefmark{1}, Roberta Chiaro\IEEEauthorrefmark{1}, Enrico Macii\IEEEauthorrefmark{2}, Massimo Poncino\IEEEauthorrefmark{1}, Daniele Jahier Pagliari\IEEEauthorrefmark{1}}%
\IEEEauthorblockA{
\IEEEauthorrefmark{1}Department of Control and Computer Engineering, Politecnico di Torino, Turin, Italy \\
\IEEEauthorrefmark{2}Interuniversity Department of Regional and Urban Studies and Planning, Politecnico di Torino, Turin, Italy}%
\IEEEauthorblockA{Email: name.first\_surname@polito.it}%
}

\maketitle

\copyrightnotice

\begin{abstract}
Collaborative Inference (CI) optimizes the latency and energy consumption of deep learning inference through the inter-operation of edge and cloud devices. Albeit beneficial for other tasks, CI has never been applied to the sequence-to-sequence mapping problem at the heart of Neural Machine Translation (NMT). In this work, we address the specific issues of collaborative NMT, such as estimating the latency required to generate the (unknown) output sequence, and show how existing CI methods can be adapted to these applications. Our experiments show that CI can reduce the latency of NMT by up to 44\% compared to a non-collaborative approach.
\end{abstract}


\section{Introduction and Related Works}
Deep learning (DL) obtains outstanding results in many Artificial Intelligence (AI) tasks that are relevant for embedded systems, such as computer vision and natural language processing (NLP).
In order to deploy DL models on embedded devices, research and industry are increasingly resorting to
Collaborative Inference (CI)~\cite{Chen2019a,daghero2020energy,Kang2017} a paradigm that combines edge and cloud computing, in an attempt to improve performance and energy efficiency. In a CI system, deep learning inference executions are distributed among a set of collaborating edge and cloud devices, with policies based on their relative compute speeds and on their current state (workload, connection speed, etc). 

Seminal works in this field mainly targeted Convolutional Neural Networks (CNNs) for computer vision~\cite{Kang2017,eshratifar2018jointdnn,Eshratifar2019,Li2019a,Huang2019}.
One of the earliest approaches is~\cite{Kang2017}, which partitions a CNN execution layer-wise between edge and cloud devices, trying to minimize its latency or energy consumption. The underlying principle is that feature tensor sizes tend to shrink for deeper layers in a CNN. Therefore, computing a few layers at the edge reduces the amount of data that needs to be sent to the cloud, and consequently the time/energy costs for transmission, possibly yielding a lower overall cost compared to pure edge and pure cloud processing. The optimal split point is adapted at runtime based on the connection latency and bandwidth, and the load of the cloud server.
This approach is extended in~\cite{Eshratifar2019}, where additionally the CNN is modified to favor partitioned execution, by inserting
layers that compress and decompress the feature maps respectively before and after transmitting them to the cloud. 
Further layer-wise partitioning approaches for feed-forward networks are found in~\cite{eshratifar2018jointdnn}, which manages multiple partition points for cases where tensor sizes are not monotonically decreasing (e.g. autoencoders), and in~\cite{Li2019a} which combines partitioning with an inference early-stopping mechanism for additional speed-ups. The authors of~\cite{Huang2019} extend these concepts to more than two offloading levels (e.g. end-device, edge gateway and cloud).

More recently, the CI paradigm has also been applied in~\cite{JahierPagliari2019,JahierPagliari2020a} to the processing of variable-length input sequences using Recurrent Neural Networks (RNNs). 
\cite{JahierPagliari2020b} extended the approach to more than two offloading levels and multiple devices in each level. 
Both works, however, focused solely on \textit{sequence-to-class} problems, such as text classification and search. In contrast, Neural Machine Translation (NMT), one of the most important DL-based tasks for smart embedded devices~\cite{googleblog}, belongs to the family of so-called \textit{sequence-to-sequence} (seq2seq) problems, where both inputs and outputs are variable-length sequences.
Previous works have shown that, when dealing with variable-length sequences, the optimal (edge or cloud) device to execute an inference depends strongly on the sequence length, which influences the computational costs~\cite{JahierPagliari2020a,JahierPagliari2020b}. However, while for sequence-to-class problems the length of the input sequence is always known beforehand, in case of a seq2seq task, computation costs also depend on the (unknown) length of the output.

In this work, we address this problem for the specific case of NMT. We show that low-cost regression models can efficiently predict the length of an output translation given the length of the input sentence, thus enabling the successful application of CI techniques. With experiments on 3 datasets and 3 DL models, based on both RNNs and Transformers~\cite{Vaswani2017}, we show that our proposed Collaborative-NMT (C-NMT) can reduce the average inference latency by up to 44\% compared to purely edge-based and cloud-based approaches, and by up to 21\% compared to a ``naive'' approach that does not account for the output length. To the best of our knowledge, ours is the first work applying CI to a seq2seq problem; we are also the first to study Transformer models from the point of view of CI.

\section{Proposed C-NMT Framework}\label{sec:method}
In this work, we propose C-NMT, a CI strategy aimed at optimizing the latency of NMT, one of the most relevant seq2seq tasks.
We build upon the work of~\cite{JahierPagliari2020a,JahierPagliari2020b}, where two key peculiarities of (sequence-to-class) NLP tasks that influence the optimal CI decisions are highlighted.
First, input/output sizes are small: encoding a sentence with the dictionary index of each word does not require more than 2 bytes per word. Thus, differently from CNNs, intermediate tensors tend to be larger than inputs, which means that \textit{partitioning} the execution between edge and cloud is not beneficial, as it doesn't help reducing data transmission costs. Instead, the optimal strategy consists in mapping an \textit{entire} inference either to edge or cloud.
Second, the \textit{length} of the processed inputs is a key parameter to be taken into account when deciding whether to run an inference at the edge or in the cloud, as it strongly influences the compute time.
In Section~\ref{sec:seq2seq} we analyze how these observations extend to NMT
and the novel challenges introduced by this task
, showing in particular that estimating the computational complexity of a translation is more complicated due to the fact that the length of the output sequence is unknown.
In Section~\ref{sec:cnmt} we then propose a simple yet effective way to solve these challenges.

\subsection{CI for Seq2Seq Deep Learning Models}\label{sec:seq2seq}

\begin{figure}[t]
  \centering
  \includegraphics[width=.96\columnwidth]{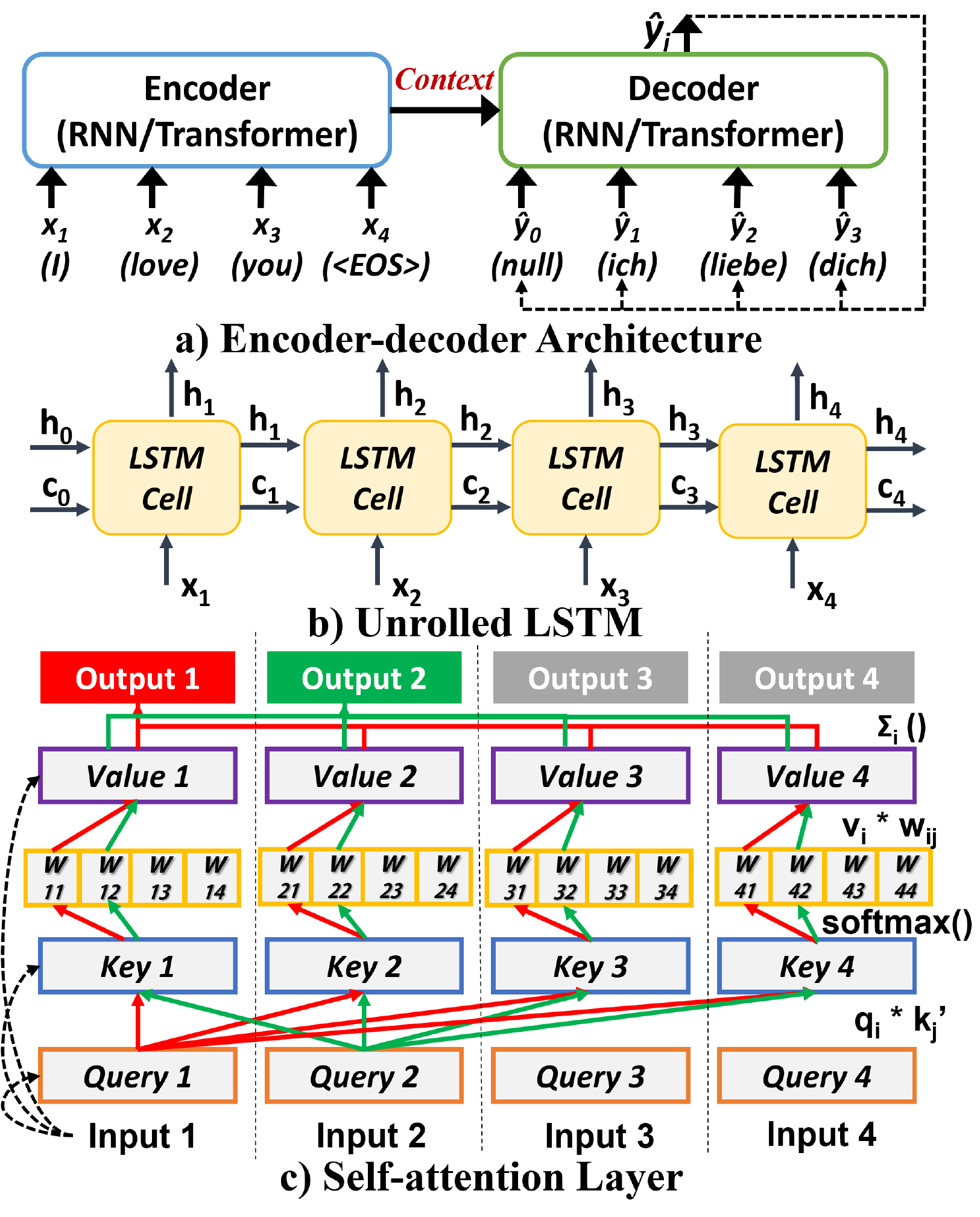}
  \caption{Encoder/decoder for seq2seq mapping and key layers.}
  \label{fig:networks}
\end{figure}

Fig.~\ref{fig:networks}a shows the most common architecture for seq2seq problems, the so called encoder/decoder. The system is composed of two separate neural networks: the \textit{encoder} (blue block) processes the variable-length input $X = x^{<1>},...x^{<N>}$, (e.g. a sentence in English) and converts it into a fixed-size, high-dimensional vector representation, called \textit{context}.
The end of the input sequence is signaled to the encoder by a special $<$EOS$>$ symbol.
The context is then fed to the \textit{decoder} (green block), whose goal is to produce the output sequence $Y = y^{<1>},...,y^{<M>}$ (e.g. the translation of the input in German). Notice that, in general, $N \ne M$. More specifically, so-called \textit{autoregressive} decoding is used in NMT, where the decoder iteratively takes as input a partially translated sentence (initially null), together with the encoder's context, and predicts the next token in the translation.

State-of-the-art models for implementing encoders and decoders are RNNs and Transformers. Here, we briefly discuss them from a computational standpoint, leaving out the details of their functionality, which can be found in~\cite{Goodfellow2016,Vaswani2017}.

RNNs are composed of one or more \textit{cells}, such as the Long-Short Term Memory (LSTM), that perform the same set of operations on each step of the input sequence, as shown in Fig.~\ref{fig:networks}b. Each step requires the output of the previous one, i.e. the hidden and cell state vectors ($\mathbf{h_i}$ and $\mathbf{c_i}$). The last cell state is used as context in encoders, whereas for decoding, hidden states are further processed with one or more fully-connected layers and softmax activations to produce word probabilities. As analyzed in~\cite{JahierPagliari2020a,JahierPagliari2020b}, the data dependency among subsequent steps makes the inference time of RNNs \textit{linearly dependent} on the processed sequence length.
This highlights a key problem of CI for RNN-based NMT. That is, estimating the \textit{total} execution time of both encoder and decoder is key to perform correct edge/cloud mapping decisions. However, while the compute time of the encoder linearly depends on the (known) input sentence length $N$, the \textit{decoder} RNN's execution time depends on the unknown (prior to the completion of the translation) output length $M$.

A similar issue arises also for Transformers. These models include several layers, but the most computationally critical is \textit{self-attention}~\cite{Vaswani2017,ivanov2020data}, shown in Fig.~\ref{fig:networks}c. For each input element, this layer generates three vectors called \textit{query} ($\mathbf{q}_i$), \textit{key} ($\mathbf{k}_i$) and \textit{value} ($\mathbf{v}_i$) through learned linear mappings, omitted in the figure for space reasons. The scalar product of each query with \textit{all} keys, followed by a softmax, produces the so-called \textit{attention weights} $\mathbf{w}_{ji}$. Finally, the $i$-th output is generated by summing together all $\mathbf{v}_j$s, each weighted by the corresponding $\mathbf{w}_{ji}$. In the figure, the flow of operations to generate the first two outputs is shown by red and green arrows respectively.
State-of-the-art transformers combine multiple of such structures (so-called attention \textit{heads}) for higher accuracy.
As for RNNs, transformer encoders typically use the output corresponding to the last (or first) input, further processed by fully-connected layers, as context.

\begin{figure}[t]
  \centering
\includegraphics[width=.99\columnwidth]{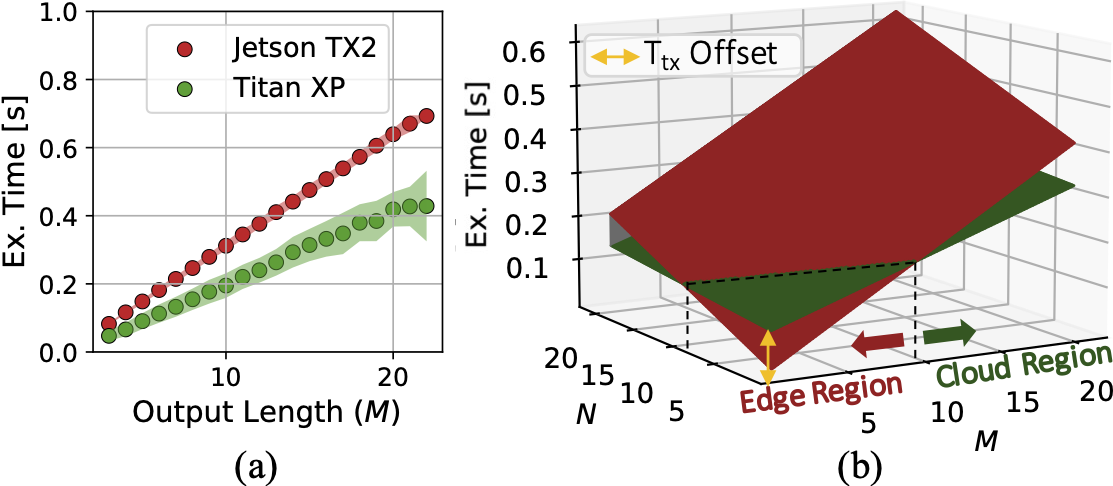}
  \caption{(a) Linear dependency of the total inference time on the output length for a Transformer. Scores of a linear fit: Jetson $R^2 = 0.99$, $MSE = 0.13ms$, Titan $R^2 = 0.85$, $MSE=1.2ms$.  (b) General principle of C-NMT.}\label{fig:method}
\end{figure}

The complexity of self-attention is \textit{quadratic} in the input length due to query-key products; however, differently from RNNs, the processing of different sequence elements can be parallelized~\cite{ivanov2020data}. Consequently, for relatively short input sequences ($<100$ tokens) and considering a highly parallel platform (e.g., an embedded GPU) we found that the inference time of Transformer encoders is approximately \textit{constant} w.r.t. $N$. In contrast, autoregressive decoding, which is implemented in Transformers with \textit{masked} attention~\cite{Vaswani2017}, imposes a strict dependency among subsequent tokens, i.e., the $i$-th predicted word is needed as input for predicting the $(i+1)$-th, limiting parallelization. In practice, the execution of the decoder has to be repeated $M$ times, which: 1) makes it significantly slower than the encoder and 2) makes the \textit{total} translation time once again linearly dependent on the output length $M$. This is clearly shown in Fig.~\ref{fig:method}a, which reports the total translation time of a Transformer as a function of $M$, for an embedded (red) and a cloud (green) GPU. The model, dataset and devices are detailed in Sec.~\ref{sec:results}.  Dots represent the average execution time for all outputs of the same length in the dataset, while colored bands represent standard deviations.

\subsection{Linear N-to-M Mapping}\label{sec:cnmt}

Fig.~\ref{fig:method}b shows visually the idea behind C-NMT. Given the analysis of Section~\ref{sec:seq2seq}, the dependency of the compute time of an NMT model on the input/output lengths approximately defines a plane in the $(N, M, T_{exe})$ space, possibly horizontal with respect to $N$ in case of transformers. Clearly, the slopes of this plane on the $z$-axis are smaller for a fast cloud device than for an edge one. However, running inference on the former has an additional latency cost related to input/output transmission ($T_{tx}$), which shifts up the ``cloud'' execution time, as shown by the yellow arrow. This generates an interesting trade-off from the point of view of CI since shorter input/output sequences are processed faster at the edge (\textit{Edge Region}) whereas cloud offloading becomes convenient only for longer input/outputs (\textit{Cloud Region}).
The intersection of the two planes, and therefore the optimal inference device for a given input depends on $N$ and $M$, on the relative speed of the involved devices, and on the time-varying $T_{tx}$. Mathematically, C-NMT selects the target device for inference $d_{tgt}$ as:
\begin{equation}\label{eq:opt}
d_{tgt} = \begin{cases}
d_e\text{ if }T_{exe,e}(N,M) \leq T_{tx} + T_{exe,c}(N,M)\\
d_c\text{ otherwise}\\
\end{cases}
\end{equation}
where suffixes $e$ and $c$ indicate edge and cloud respectively. Given, the compact encoding of inputs/outputs in NMT discussed above, in this work we model $T_{tx}$ as being dominated by the connection's round-trip time, and roughly dependent on $N$ and $M$. As shown also in~\cite{JahierPagliari2020a,JahierPagliari2020b}, although this is an approximation, it yields quite accurate CI decisions.
Concerning $T_{exe}$, given the analysis of Sec~\ref{sec:seq2seq}, we model it as a linear function of $N$ and $M$, that is $T_{exe,i} = \alpha_{N,i} \cdot N + \alpha_{M,i} \cdot M + \beta_{i}$, where $i \in \{e, c\}$ and $\alpha_{N/M,i}$, $\beta_{i}$ depend on the compute power of device $d_i$ and on the NN model. These parameters can be computed with a once-for-all offline characterization.

\begin{figure}[t]
  \centering
  \includegraphics[width=.33\linewidth]{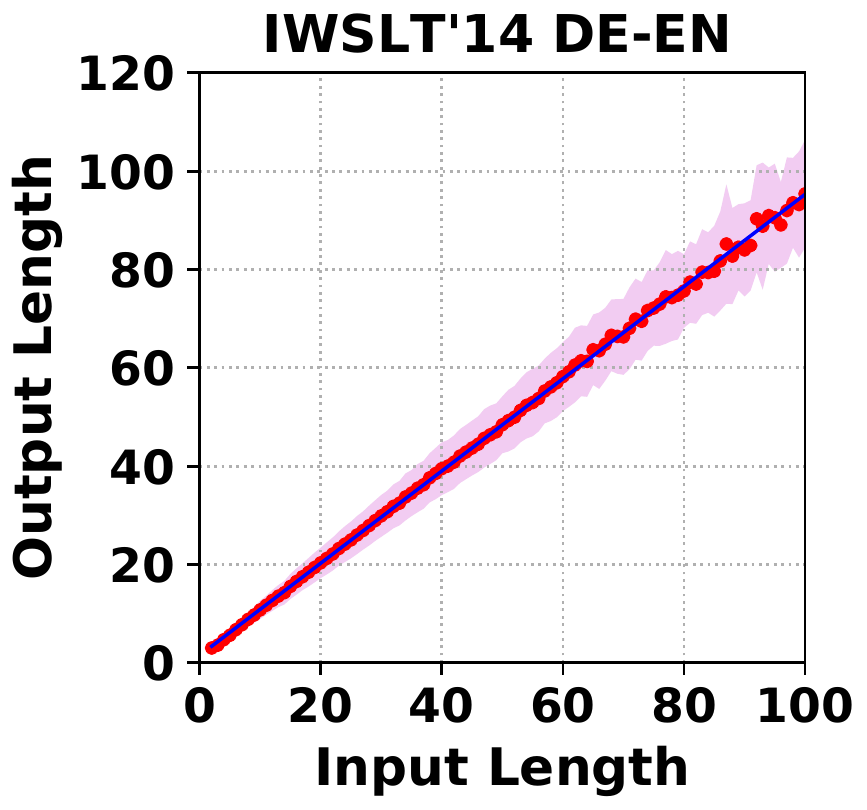}%
  \includegraphics[width=.33\linewidth]{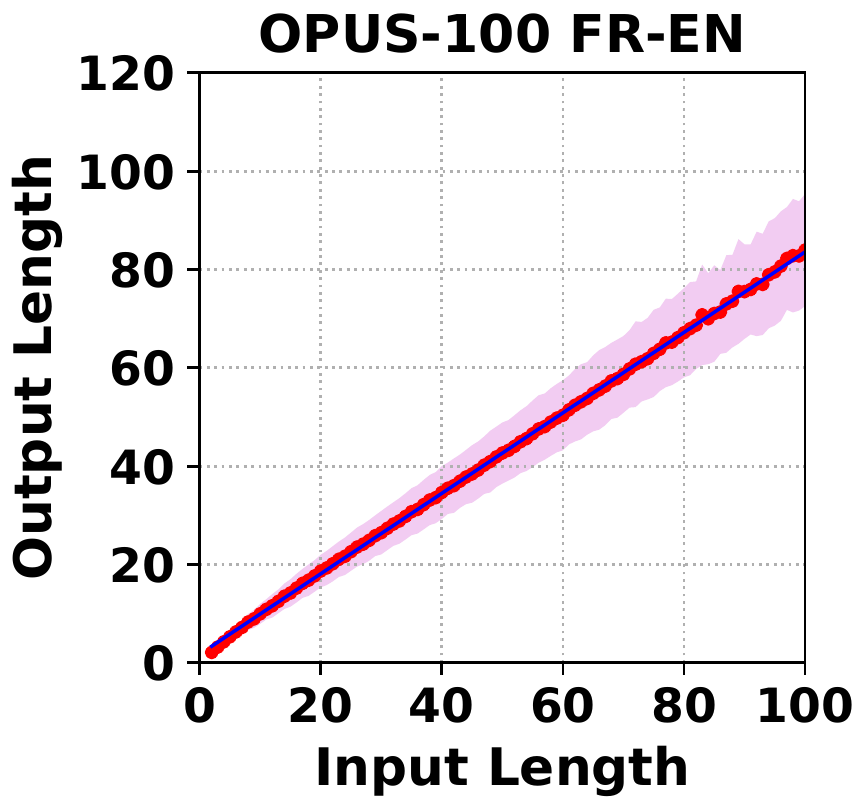}%
  \includegraphics[width=.33\linewidth]{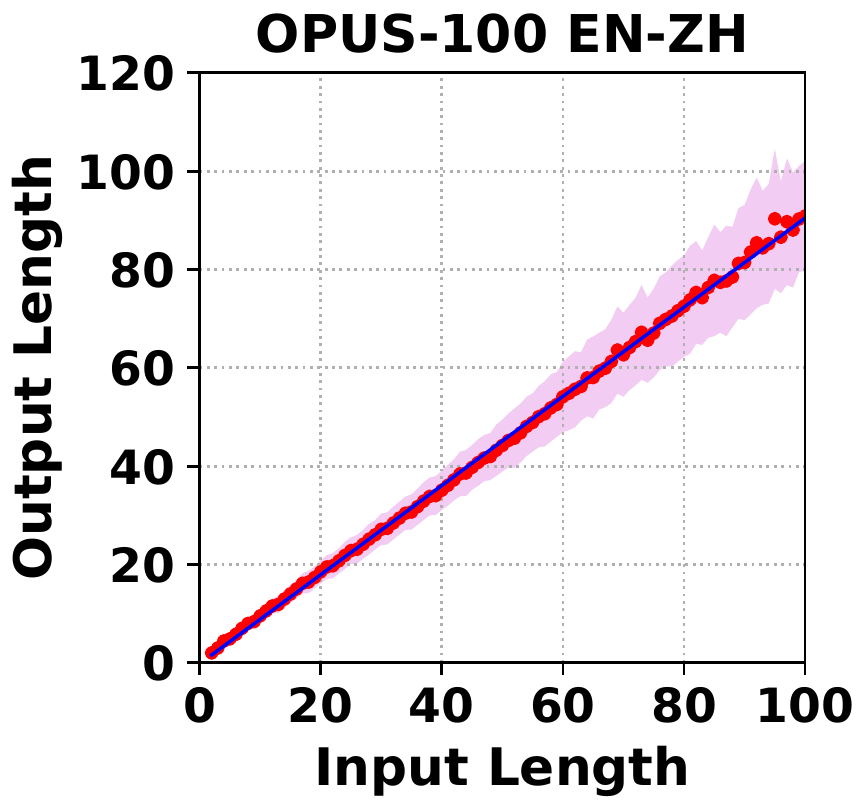}%
  \caption{Regression models for the output length estimate. IWSLT'14,DE-EN: R2-score=0.99, MSE=0.57; OPUS-100,FR-EN: R2-score=0.99, MSE=0.15; OPUS-100,EN-ZH R2-score=0.99, MSE=0.73.}
  \label{fig:in_out_example}
\end{figure}

The most critical quantity in the above equation is $M$, which only becomes known \textit{after} the completion of a translation. However, for the particular case of NMT, it is reasonable to assume that there is a correlation, to some extent, between the length of an input sentence and the one of its translation. As an example, Fig.~\ref{fig:in_out_example} shows the average $M$ for a given $N$ and the corresponding standard deviation for three different language pairs. The caption reports the excellent regression scores obtained by a simple linear model relating the two quantities. These results show that, even for very different languages, such as Chinese and English, an accurate estimate of the output length can be obtained with a simple linear $N$-to-$M$ mapping. This is the strategy used in our proposed CI system, which eventually estimates $T_{exe,i}$ as:
\begin{equation}\label{eq:t_exe}
    T_{exe,i} = \alpha_{N,i} \cdot N + \alpha_{M,i} \cdot (\gamma\cdot N + \delta)+ \beta_{i}
\end{equation}
where $\gamma$ and $\delta$ are correcting factors that only depend on the target language pair, and are independent from the device and neural network model. The need for this correction is evident from Fig~\ref{fig:in_out_example}, which clearly shows that $\gamma < 1$ is needed to account for the lower verbosity of the English language (EN) with respect to French (FR) in Fig.~\ref{fig:in_out_example}b, and of Chinese (ZH) with respect to English in Fig.~\ref{fig:in_out_example}c.

\begin{table*}
\begin{minipage}{0.3\linewidth}
    \centering
    \includegraphics[width=.95\columnwidth]{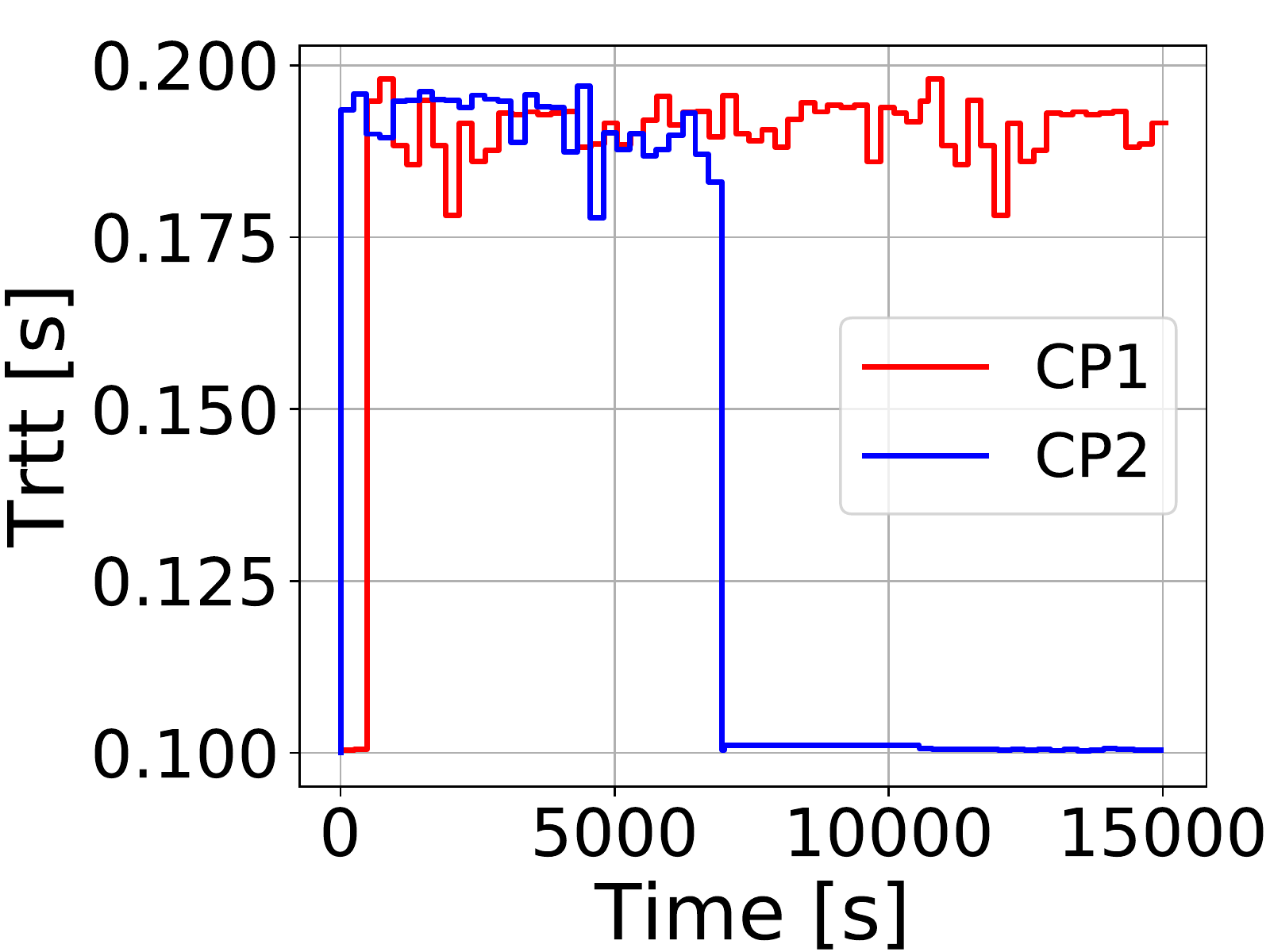}
    \captionof{figure}{Connection profiles.}
    \label{fig:cp}
\end{minipage}\begin{minipage}{0.7\linewidth}
\centering
\footnotesize
\caption{\label{tab:table-name} Execution time variation (in \%) for two different variable connection profiles.}\label{table:table_2_levels_dac}
\begin{tabular}{|c|c|c|c|c|c|c|c|}
\hline
\multirow{3}{*}{\textbf{Dataset}} & \multirow{3}{*}{\textbf{Strategy}} & \multicolumn{3}{c|}{\textbf{Connection Profile (CP) 1}} & \multicolumn{3}{c|}{\textbf{Connection Profile (CP) 2}}\\\cline{3-8}
& & \textbf{Ex. Time} & \textbf{Ex. Time} & \textbf{Ex. Time} & \textbf{Ex. Time} & \textbf{Ex. Time} & \textbf{Ex. Time}\\
& & \textbf{vs. GW} & \textbf{vs. Server} & \textbf{vs. Oracle} & \textbf{vs. GW} & \textbf{vs. Server} & \textbf{vs. Oracle}  \\\hline
\multirow{2}{*}{DE-EN} & Naive & +11.74 & -4.78 & +29.17 & -16.16 & -5.28 & +13.25\\
& C-NMT & \textbf{-13.55} & \textbf{-26.15} & \textbf{+0.11} & \textbf{-24.34} & \textbf{-17.65} & \textbf{+0.15}\\\hline
\multirow{2}{*}{FR-EN} & Naive & -5.74 & -40.80 & +8.03 & -7.15 & -32.13 & +15.46\\
& C-NMT & \textbf{-12.29} & \textbf{-44.32} & \textbf{+1.24} & \textbf{-18.00} & \textbf{-41.06} & \textbf{+1.13}\\\hline
\multirow{2}{*}{EN-ZH} & Naive & -17.11 & -8.08 & +15.49 & \textbf{-36.31} & \textbf{-10.41} & \textbf{+8.51}\\
& C-NMT & \textbf{-21.17} & \textbf{-12.46} & \textbf{+9.83} & -35.66 & -10.58 & +8.77\\\hline
\end{tabular}
\end{minipage}
\end{table*}

\subsection{Implementation Details}

After the offline characterization of the target NN model, the C-NMT decision has negligible overheads, as it simply consists of evaluating (\ref{eq:t_exe}) and (\ref{eq:opt}). For what concerns $T_{tx}$, although this quantity is roughly independent of $N$ and $M$, it still changes over time due to the variability of the edge-cloud connection signal quality or data traffic. As in~\cite{JahierPagliari2020b}, we attach \textit{timestamps} to each inference request/response sent to/from the cloud to obtain a recent estimate of $T_{tx}$. However, on end-nodes (e.g., smartphones), translation tasks are typically performed sporadically, rendering the timestamp mechanism ineffective. For this reason, we consider a system where the edge device is a \textit{gateway} which aggregates the requests of multiple end-nodes and therefore can be assumed to be almost continuously fed with inference requests. The C-NMT decision then becomes whether to perform inference locally at the gateway or in a more powerful cloud server.

\section{Experimental Results}\label{sec:results}

We assess the effectiveness of C-NMT considering: i) an edge gateway (GW) made of an NVIDIA Jetson TX2, including a Pascal GPU with 256 CUDA cores, and ii) a cloud server equipped with a Dual Intel Xeon E5-2630@2.40GHz, 128GB RAM and an NVIDIA Titan XP GPU. Both devices run Linux and perform inference with PyTorch.
For repeatability, the network connection between the devices is simulated using 2 real round-trip-time ($T_{rtt}$) profiles taken RIPE Atlas~\cite{atlas}, and assuming a constant and symmetric bandwidth of 100Mbps. The simulation time vs $T_{rtt}$ traces are shown in Fig~\ref{fig:cp}, and refer to the following RIPE Atlas query: meas\_id: 1437285; probe\_id: 6222; Date $=$ 03/05/2018; Time $=$ 3-7 p.m. (CP1), 7:30-12:30 a.m. (CP2).

The experiment consists in sending 100K translation requests to the GW, which uses C-NMT to decide, for each input, whether to process it locally or offload it to the cloud.
The $T_{exe}$ model of (\ref{eq:t_exe}) is fitted on the result of 10k inferences per device, with inputs not included in the 100k set.

We repeat the experiment for 3 different NMT architectures and datasets: i) A 2-layer BiLSTM model~\cite{OpenNMT} with a hidden size of 500, tested on the IWSLT'14 German-English (DE-EN) corpus~\cite{iwslt}; ii) A single-layer Gated Recurrent Unit (GRU) RNN~\cite{Punchline} with hidden size 256, tested on the OPUS-100 French-English (FR-EN) corpus~\cite{zhang2020improving}; iii) The ``MarianMT'' attention-based Transformer~\cite{wolf2019huggingface_marianMT} tested on the OPUS-100 English-Chinese (EN-ZH) corpus~\cite{zhang2020improving}.
For each dataset, the correcting factors $\gamma$ and $\delta$ in (\ref{eq:t_exe}) are computed on the \textit{ground-truth} $(N, M_{real})$ pairs in the corpus, where $M_{real}$ may in general differ from the output length $M$ produced by the NMT model. Further, when computing $\gamma$ and $\delta$, we remove outliers (e.g., wrongly matched sentence pairs) following the pre-filtering rules described in~\cite{banon2020paracrawl_paper_cleaning}.

Table~\ref{table:table_2_levels_dac} reports the obtained results. As baselines for comparison, we consider the 2 single-device approaches, i.e., the scenarios in which \textit{all} 100k inputs are processed in the GW or in the server. Moreover, to have an ideal lower bound on latency, we consider an \textit{Oracle} policy capable of always selecting the fastest inference device, without being affected by the sources of sub-optimality of C-NMT, such as the imperfect N-to-M regression, the linear $T_{exe}$ model, the outdated $T_{tx}$ estimates, etc. Lastly, we compare C-NMT against a CI strategy that uses the same mapping policy, but simply assumes $M$ equal to the average output length of the reference dataset when estimating $T_{exe}$. We call this approach \textit{Naive}, and we use it to show the positive impact of N-to-M mapping in C-NMT. The results in the table are reported as percentage variations in the total ex. time for the 100k inferences, with respect to the 3 baselines (single devices and Oracle), where negative/positive numbers indicate an ex. time reduction/increase respectively.

The results show that, by mapping each translation either to the GW or to the server based on the input and (predicted) output lengths, C-NMT is able to significantly reduce the execution time compared to purely edge-based and cloud-based approaches. The total time reduction is up to 26\%, 44\% and 36\% respectively for DE-EN, FR-EN and EN-ZH translations. As expected, the benefit of C-NMT w.r.t to a cloud based approach is larger with the first connection profile (CP1), which is slower on average, and therefore makes cloud offloading sub-optimal except for very long sentences, most of the time. The opposite reasoning applies to the comparison with a pure edge computing approach.

Also expectedly, the overhead of C-NMT w.r.t. an Oracle policy are larger for the EN-ZH transformer than for the two RNNs. This is because, as analyzed in Sec.~\ref{sec:seq2seq}, decoding dominates the total latency of Transformers-based NMT on GPU platforms. Therefore, the $T_{exe}$ estimate for this type of model relies more heavily on the unknown $M$, thus suffering more from the approximated N-to-M mapping.
Lastly, C-NMT is significantly more effective than the Naive approach (up to 21\% larger ex. time reduction for the DE-EN dataset), except for EN-ZH translation with CP2, where the two approaches achieve very similar results.

\section{Conclusions}
We have presented C-NMT, the first collaborative inference framework for NMT applications based on deep learning. We have tested our approach on RNNs and Transformers, the two state-of-the-art architectures for this type of problem, demonstrating significant execution time reductions (up to 44\%) with respect to any static mapping solution.
Future works will focus on the study of more advanced output length estimation methods.

\bibliographystyle{IEEEtran}

\end{document}